\documentclass[11pt]{article}

\usepackage{ARR2023/acl}

\usepackage{times}
\usepackage{latexsym}

\usepackage[T1]{fontenc}

\usepackage[utf8]{inputenc}

\usepackage{microtype}
\usepackage{inconsolata}

\usepackage{textcomp}
\usepackage[utf8]{inputenc}
\usepackage{amsmath,amssymb}
\usepackage[ruled]{algorithm2e}
\usepackage{algorithmic}
\usepackage{graphicx}
\usepackage{array}
\usepackage{booktabs}
\usepackage{float}
\usepackage{caption}
\usepackage{natbib}
\usepackage{cite}
\usepackage{multirow}
\usepackage{comment}
\usepackage{mathtools}

\usepackage{amsfonts}
\usepackage{url}
\setcitestyle{authoryear}

\DeclareMathOperator*{\argmax}{\mathrm{argmax}}

\usepackage{url,hyperref}
\usepackage{color}

\title{Prompt-Based Length Controlled Generation with Multiple Control Types}

\author{Renlong Jie \\ Northwestern Polytechnical University
        \And  Xiaojun Meng, Lifeng Shang, Xin Jiang, Qun Liu \\ Huawei Noah's Ark Lab
        }

\author{Renlong Jie\textsuperscript{\rm 1*}, Xiaojun Meng\textsuperscript{\rm 2}, Lifeng Shang\textsuperscript{\rm 2}, Xin Jiang\textsuperscript{\rm 2}, Qun Liu\textsuperscript{\rm 2} 
    \\\textsuperscript{\rm 1}Northwestern Polytechnical University, \textsuperscript{\rm 2}Huawei Noah's Ark Lab\\
    jierenlong@nwpu.edu.cn, \{xiaojun.meng, Lifeng.Shang, Jiang.Xin, qun.liu\}@huawei.com
        }

\begin{document}

\maketitle
\renewcommand{\thefootnote}{\fnsymbol{footnote}} 
\footnotetext{*Work is done as a postdoctoral research fellow at Noah's Ark Lab, Huawei.}

\begin{abstract}

Large language models (LLMs) have attracted great attention given their strong performance on a wide range of NLP tasks. In practice, users often expect generated texts to fall within a specific length range, making length controlled generation an important topic, especially for GPT-style models. 
Existing length control methods mostly focus on a simple control type of ``equal to'' a target length. Different from them, we propose a prompt-based method to achieve length controlled generation under different control types with high accuracy. In particular, we adopt reinforcement learning (RL) and sample filtering with the reward signal given by rule-based reward models, which enhances the length control ability of models by rewarding outputs that follow certain control instructions. 
In addition, we introduce a standard prompt extractor to parse arbitrary users' input into standard control instructions. Experiments show that our method significantly improves the accuracy of prompt-based length control on popular summarization datasets like CNNDM and NYT under multiple control types. Moreover, both the standard prompt extractor and RL-tuned model show strong generalization to unseen control prompt templates.

\end{abstract}

\section{Introduction} \label{Sec:1}

For recent popular GPT-style models like ChatGPT and GPT-4 \citep{radford2018improving, radford2019language, liu2023summary, openai2023gpt4}, various studies have been conducted on them, and the inference efficiency and computational cost often draw concerns from the community~\citep{zhang2023complete, zhao2023survey, bubeck2023sparks}. Since its generation is in an autoregressive manner, the inference cost increases continually with the growing of decoding steps. Meanwhile, users of LLMs usually have an expected length of generated texts, no matter for writing an essay or summary, knowledge QA or dialogue generation \citep{fan2018controllable,liu2020asking,liu2022length, mirshekari2021conquest, gupta2021controlling}. Both of these two facts require the length of generation in GPT-style models can be effectively controlled. 

For pretrained language models (PLMs), the most widely applied technique for length control is prompt-based fine-tuning \citep{raffel2020exploring, goyal2022news, zhang2022latent, liu2023pre}. Taking an example of length-controlled summarization (LCS), we can prepend a prompt ``\texttt{summarize with length $l_i$:}'' to the article to be summarized in training, where $l_i$ is the number of words of reference summary. However, this process is usually performed in supervised fine-tuning (SFT), where the length control ability has to compromise with the goodness of downstream tasks. For large-scale models like GPT-3, the length controlled ability can be somewhat activated by in-context learning without any fine-tuning~\citep{brown2020language, chowdhery2022palm, dong2022survey}. However, it relies on the size and power of the pre-trained foundation models to achieve good performance. 
For other methods like RLHF 
~\citep{christiano2017deep, stiennon2020learning, ouyang2022training}, 
it is expensive to manually label whether the generated length meets the requirement given in instruction prompts. 
Meanwhile, automatic reward labels can not be straight forwardly obtained, as the control instructions can be arbitrarily integrated into user utterances. 

Generally, there are many other length control methods such as GOLC, LenAtten and LAAM \citep{liu2018controlling, takase2019positional, makino2019global, yu2021lenatten, liu2022length}. However, these methods are not particularly designed for PLMs, thus architecture-specific designs on training mechanisms are usually needed. Moreover, they often focus on the setting of equalling to a certain length, generally not adapt to other control types such as greater/smaller than a value, or between two values, \textit{etc}. Meanwhile, they can not handle diverse expressions of control instructions from users. Therefore, how to effectively connect diverse control instructions from users to the length of generated text for PLMs is still an open question.

In this paper, we introduce a novel method that applies prompt-based fine-tuning with reinforcement learning to improve the performance of length controlled generation, which is capable to handle multiple types of length control at the same time. Our main contributions are:
\vspace{-0.5em}
\begin{itemize}
\item We design a rule-based reward model for multiple control types other than traditional ``equal to'' control type, which can provide accurate reward values for both reinforcement fine-tuning and inference of PLMs.
\vspace{-0.5em}
\item We introduce an independent standard prompt extractors (SPE) to parse the length control instructions from diverse user inputs to standard control prompts (SCP), which is necessary for rule-based reward and show strong generalization power. 
\vspace{-0.5em}
\item We apply a Proximal Policy Optimization (PPO) algorithm with a modified state space to fine-tune GPT models for enhancing their length control ability. Two modes including (a) SCP + rule-based reward; (b) SCP + model-based reward are introduced. 
\vspace{-0.5em}
\item Experiments show that by applying reinforcement fine-tuning and sample filtering, the length-control errors can be significantly reduced from the baseline prompt-based method. Moreover, the method show strong generalization to unseen prompt templates.
\end{itemize}

\section{Related work} \label{Sec:2}

\subsection{Reinforcement learning for text generation.} \label{Sec:2.1}

Reinforcement learning (RL) \citep{kaelbling1996reinforcement} has been widely applied to improve text generation performance, including summarization \citep{stiennon2020learning, paulusdeep2018}, question generation \citep{pang2021text}, and dialogue generation \citep{li2016deep, zhou2017end, jaques2020human}. In general, we can consider the generative model as the policy network and optimize its parameters for achieving higher reward from the environment \citep{paulusdeep2018, wang2022deep}. Human feedback is one of the most known strategies to get the reward, which is shown to be more effective than using some automatic metrics, such as rouge scores in text generation \citep{christiano2017deep, stiennon2020learning, wu2021recursively}. Existing study \citep{ramamurthy2022reinforcement} also shows that RL techniques are generally better than supervised methods at aligning language models to human preferences. It is known that Reinforcement learning from Human Feedback (RLHF) plays a key role in the success of autoregressive LLMs like InstructGPT~\citep{ouyang2022training}, which uses human feedbacks to train a reward model for PPO \citep{schulman2017proximal}. 

\subsection{Length control for text generation} \label{Sec:2.2}

Length control is an important ability for text generation, especially for tasks with a large variance of output length, such as summarizing texts using a desired range of number of words/tokens. 
In this work, we particularly focus on the text summarization task,  which is the most concerned task for length controllable text generation. 

Early work~\citep{fan2018controllable} on controlling lengths in abstractive summarization quantifies summary length into discrete bins, and expands the input vocabulary with special tokens to indicate the length bins of the ground-truth summary during training. 
\citep{liu2018controlling} extends a convolutional sequence to sequence model to control the length of summarization. To generate summaries of any desired length, a length constrain factor is added to each convolutional block of the initial layer. \citep{takase2019positional} proposes an extension of a sinusoidal positional encoding to enable neural encoder-decoder model to generate a text of any desired length. GOLC ~ \citep{makino2019global} dedicates to increase the probabilities of generating a high quality summary within a desired length by using minimum risk training. LenAtten \citep{yu2021lenatten} introduces a length attention unit to break the trade-off between length controllability and summary quality. LAAM \citep{liu2022length} modifies the attention matrix based on length-budget dynamically during the decoding process. Generally, we notice that existing length control approaches can not be directly applied for control targets other than ``equal to'' a certain length, and are in lack of focusing on prompt-based method for the most recent trend of GPT-style LLMs. 

\section{Method} \label{Sec:3}

We aims to develop the length-controlled generation methods for GPT-style PLMs, especially for the cases with multiple control types.
We first introduce the whole architecture in Section~\ref{Sec:3.1} and then discuss each component of it.

\subsection{Model Architecture} \label{Sec:3.1}

\begin{figure}[t]
\begin{center}
\includegraphics[width=1.0\linewidth]{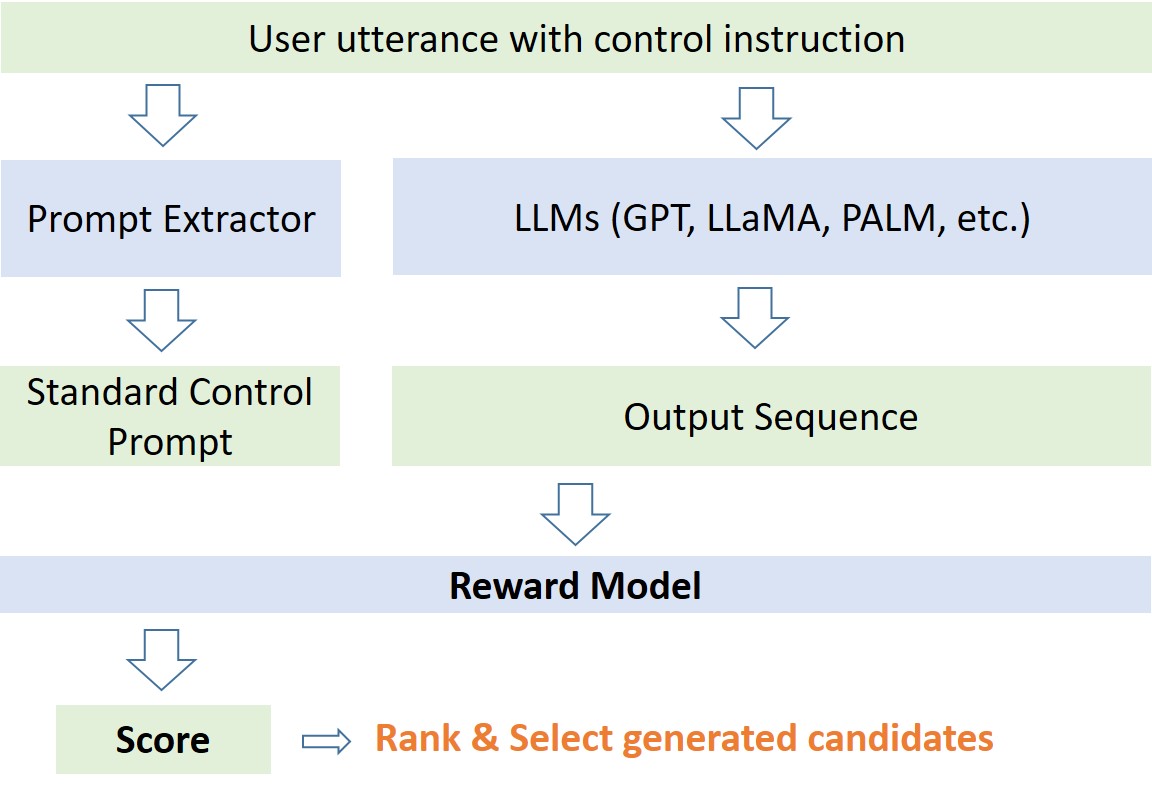}
 \caption{Overview of the model architecture. In training stage, the scores given by the reward model are used for the reinforcement learning method. In inference stage, the scores are applied for ranking and selecting the output sequences generated by PLM/LLMs.} \label{Fig:1}
\end{center}
\vspace{-0.10em}
\end{figure}
Our model architecture is presented in Figure~\ref{Fig:1}. The original user utterances may include the control instruction on length constraint, which differs from factual and semantic information in terms of that the length control can be easily checked by rule-based methods. For instance, if we can understand user intention on length constraint, we can set up the rule for ranking and selecting generated candidates.
Therefore, we introduce a standard prompt extractor (SPE) (See Section~\ref{Sec:3.2}) to parse the information of length constraint from user utterance and thus generate a standard control prompt. This standard prompt includes different types of length constraint and can be applied for rule-based inference and evaluation. 

As Figure~\ref{Fig:1} shows, the user utterance is first passed through both the SPE and PLM/LLMs like GPT-family~\citep{brown2020language, openai2023gpt4}, PALM~\citep{chowdhery2022palm}, LLaMA~\citep{touvron2023llama}, Pangu~\citep{ren2023pangusigma}, Ernie~\citep{sun2020ernie}, etc. PLMs are the core modules that generate an output sequence according to the user utterance, and SPE outputs a standard control prompt (SCP) that includes user intention on the control type and target lengths. Secondly, the reward model takes both the SCP and generated sequence as input, and outputs a score to evaluate how well the generated sequence meets the requirement of length control instruction (See Section~\ref{Sec:3.3}). Finally, this score can be applied as the reward signal in reinforcement learning method to fine-tune PLMs (See Section~\ref{Sec:3.4}), or be applied as a filtering rule to rank and select the generated sequences in inference (see Section~\ref{Sec:3.5}).



\subsection{Reward model} \label{Sec:3.3}

\begin{table}[t]
  \centering{\footnotesize
    \begin{tabular}{p{9em}p{11em}} 
    \toprule
    \textbf{Standard Control Prompt} & \multicolumn{1}{c}{\textbf{Reward}} \\
    \midrule
    more than $L_{t}$ &  $-\max(0, L_{t}-L_{g})$ \\
    less than $L_{t}$ & $-\max(0, -L_{t}+L_{g})$ \\
    equal to $L_{t}$ & $-|L_{t}-L_{g}|$ \\
    between $L_L$ and $L_U$ &  $-(\max(0, L_L-L_{g}) + \max(0, L_{g} - L_U))$  \\
    \bottomrule
    \end{tabular}%
      \caption{Reward function for each Standard Control Prompt (SCP). We provide the plots of these functions in Appendix A.1}
  \label{tab:1}}%
  \vspace{-0.10em}
\end{table}
To evaluate whether the generated text follows the length control instruction, we introduce a reward model to score the generated sequences according to the required length from the user's input. This score can be used as a reward for fine-tuning existing PLMs by leveraging reinforcement learning, or be used to rank and select the candidates generated by PLMs. 
In this study, we design a \textbf{rule-based reward model}, which takes the actual length of the output sequence and target values as the inputs, and calculate the rewards using the reward functions depending on the type of SCPs, as is shown in Table~\ref{tab:1}, where $L_t$, $L_L$, $L_U$ and $L_g$ refer to the target length, the lower-bound length, the upper-bound length and the actual generated length, respectively.
The type of SCPs and target lengths are parsed from user's input as is shown in Figure ~\ref{Fig:1}. The rule-based method provides the accurate feedback on whether the output meets the requirement of length given by SCPs, while the latency is almost negligible compared with using a neural model (\textit{e.g.,} BERT or GPT) for scoring. However, it relies on extracting exact standard control information from the user's input. We also discuss the model-based reward models in Appendix~\ref{Appendix:stan_prompt}, which are generally outperformed by rule-based ones. 

\subsection{Standard Prompt Extractor} \label{Sec:3.2}
\begin{figure*}[th]
\begin{center}
 \includegraphics[width=1.0\linewidth]{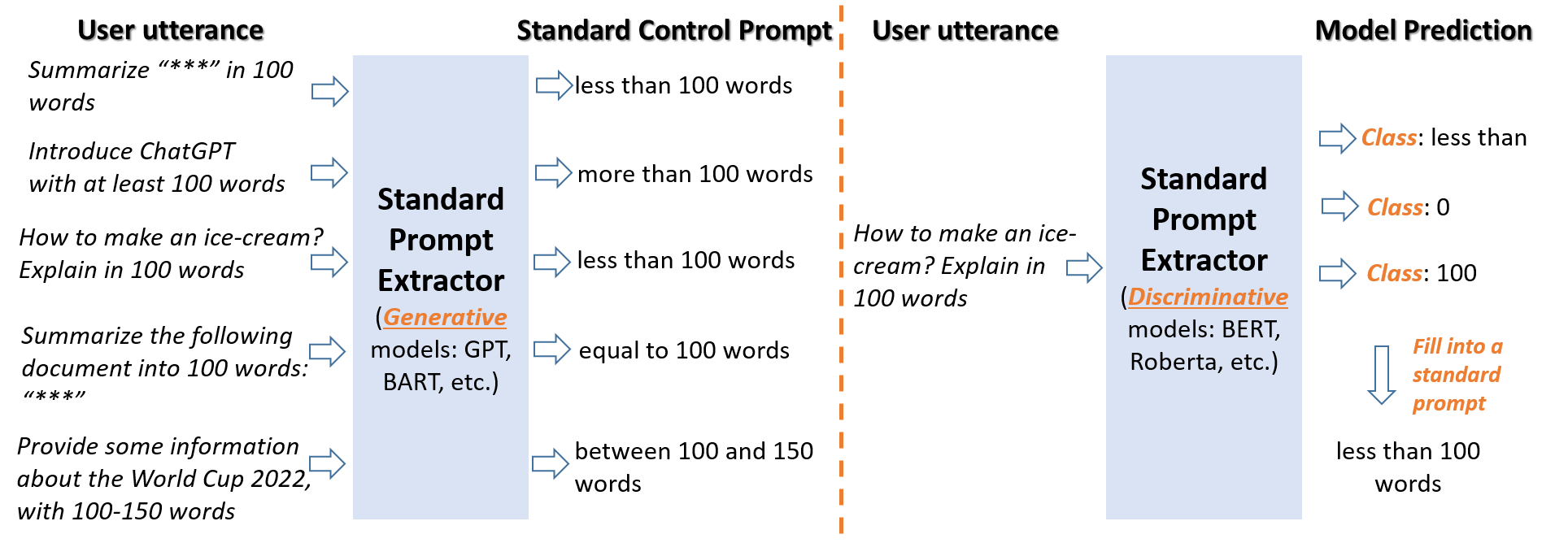}
 \caption{The demonstration of Standard Prompt Extractor (SPE). The generative type of models are trained to output the standard control prompts (SCPs) directly (left), while the discriminative type of models are trained to predict the type of each control instruction, as well as the requested number of lengths from user utterance, such as the minimum value and the maximum value (right).} \label{Fig:2}
\end{center}
\end{figure*}

To get SCPs for applying rule-based reward model to score the generated sequences in RL and sample filtering, we introduce Standard prompt extractor (SPE), which takes a user's input, and outputs standard control prompt (SCP) if exists. This standard prompt consists of a basic description of what length constraint should be satisfied. We design two types of SCPs as shown in Figure~\ref{Fig:2}.
In particular, this prompt extractor can be a \textbf{generative model} such as GPT, in which case the extractor is trained directly to generate the full text of SCP as shown in Figure~\ref{Fig:2} (left). Then we can easily get $L_t$, $L_U$ and $L_L$ of Table~\ref{tab:1} from this generated control text. On the other hand, we can also use a \textbf{discriminative model} such as BERT, as the prompt extractor, in which case it is required to predict the type of SCP and the target numbers involved, as shown in Figure~\ref{Fig:2} (right).
In this case, we prepend three \texttt{[CLS]} tokens to the utterance. 
Three linear projection layers with different output sizes (\textit{i.e.,} number of types of control instruction as in the left column of Table~\ref{Fig:1}, number of possible minimum values, number of possible maximum value) map the three top vectors of \texttt{[CLS]} tokens to fill in the type, minimum value and maximum value of a standard prompt template. Therefore, we have three classification objectives for predicting the ground truth of SCP. Note that we can indeed use only the minimum and maximum target values to fully represent the control instructions under all the four types in Table~\ref{tab:1}. For example, the minimum target value is 0 means the control type of ``\texttt{smaller than}'' the maximum target value.
Since this setting has only two classification objectives, two \texttt{[CLS]} tokens and corresponding linear projection layers are introduced.

\subsection{Reinforcement Learning for length control fine-tuning} \label{Sec:3.4}

We apply a modified PPO method with actor-critic setting \citep{grondman2012survey, bahdanauactor, schulman2017proximal}. Since rewarding of the generated text length does not rely on the input article, both the reward model and critic model only take the concatenation of SCPs and generated texts as input. 
Meanwhile, as the reward for length control can only be determined when the generation ends, we can just calculate the reward using the final output. Assume $\pi_{\theta}(a|s)$ is a stochastic policy given by the PLM, where $\theta$ is the trainable parameter, $s$ is the whole input sequence, and $a$ is the finally generated sequence. Let $s'$ be the SCP. The original policy gradient (PG) applies the loss function given by Eq.~\eqref{Eq:1}:
\begin{equation}
L^{PG}(\theta) = -\hat{\mathbb{E}}_D[\log\pi_{\theta}(a|s)\hat{A}],
\label{Eq:1}
\end{equation}
where $\hat{\mathbb{E}}_D[.]$ is the empirical average over a finite batch of samples from dataset $D$. $\hat{A}$ is an estimator of the advantage function at the end of generation, showing the goodness of current policy \textit{w.r.t.} the baseline in terms of control accuracy. For the actor-critic case, we set 
$\hat{A}= R(s',a)-\hat{Q}_{\theta_{old}}(s',a)$, where $R(.)$ is the reward model, $\hat{Q}_{\theta_{old}}(s',a)$ is the expected Q value by the model of the last step. Note that the reward only depend on the standard control prompt $s'$ and the generated sequence $a$ (without the input context). As $s'$ itself is not associated with the control reward, it is hard to define a value function on it. Thus, we apply $\hat{Q}_{\theta_{old}}(s',a)$ instead of $V_{\theta_{old}}(s')$ as the baseline of the current step. The original PG empirically often leads to a large policy update and thus instability during fine-tuning. 
Therefore, we follow PPO~\citep{ schulman2017proximal} to use the probability ratio $r(\theta)=\frac{\pi_{\theta}(a|s)}{\pi_{\theta_{old}}(a|s)}$ instead of $\log\pi_{\theta}(a|s)$ in Eq.~\eqref{Eq:1}, and utilizes a clipped surrogate objective given by Eq.~\eqref{Eq:2} to stabilize the policy updates and ensure that the probability ratio term is bounded between $[1-\epsilon, 1+\epsilon]$.
\begin{equation}
\begin{split}
L^{CLIP}(\theta) &= -\hat{\mathbb{E}}_D[\min(r(\theta)\hat{A}, \text{Clip}(r(\theta), \\
&1-\epsilon, 1+\epsilon)\hat{A})].
\end{split}
\label{Eq:2}
\end{equation}
To ensure sufficient exploration, we also follow the original paper of PPO to introduce an entropy term $S=\frac{1}{n}\sum \pi_{\theta}(a|s)\log(\pi_{\theta}(a|s))$, in which the average is taken across the vocabulary dimension. In addition, a penalty for large KL divergence is added between the current and old stochastic policy distributions (i.e. $\pi_{\theta}$ and $\pi_{\theta_{old}}$).  
Therefore, the total policy loss can then be rewritten as:
\begin{equation*}
\begin{split}
    L^{CLIP+S+KL}(\theta)&=\hat{\mathbb{E}}_D[L^{CLIP}(\theta)-cS[\pi_{\theta}|(s)] \\
    &+ \beta D_{KL}(\pi_{\theta}|\pi_{\theta_{old}})],
    \end{split}
    \label{Eq:3}
\end{equation*}
where $c, \beta$ are coefficients, $D_{KL}(\pi_{\theta}|\pi_{\theta_{old}})$ is the KL-divergence between the old and current action probability distributions. To avoid the performance loss for downstream tasks, we involve an extra term of SFT loss from the same batch of labelled data on the actor's policy loss: $L^{A}(\theta) = L^{CLIP+S+KL}(\theta) + \lambda L^{SFT}(\theta)$, where $\lambda$ is a tunable hyper-parameter. Meanwhile, we optimize a value loss $L^{VF} = (Q_{\theta}(s', a) - \hat{R})^2$. More details of the algorithm are given in Appendix~\ref{Appendix:Alg}. 


\subsection{Inference \& Sample filtering} \label{Sec:3.5}

In inference, a well fine-tuned PLM is expected to directly process user inputs, and generate a text sequence following the user's intention on length control. Since the control instruction from the user inputs can be diverse in practice, our proposed prompt extractor serves as an important role to parse user inputs into SCPs to benefit the latter RL fine-tuning. Meanwhile, with the extracted type and value information from SPEs, we can apply reward models (as described in Section~\ref{Sec:3.2}) to score, rank and select from a set of generated samples in beam sampling, which is named as sample filtering in our method. Let $k=\argmax_i R(s', a_i)$, where $R$ is the reward model, $a_i$ is the $i$-th sequence in all $N$ output sequences, then a $a=a_k$ is selected to be the final output sequence. Thereafter, this selected sequence can be used for either the RL fine-tuning phase or the final inference for validating to what extent the length control ability can be achieved in existing PLMs.

\section{Experiments} \label{Sec:4}

\subsection{Experimental Setup} \label{Sec:4.1}

Although our method can work for all types of length controlled text generation tasks, we focus on summarization task in our experiments. This is because we believe summarization is the most concerned task for length controllable text generation. Meanwhile, it has standard automatic metrics and comparable benchmarks. Almost all existing works on length-controllable text-text generation focus on summarization task. For Q\&A and dialog tasks, the suitable length of answering highly depends on the questions, while strict length control towards randomly sampled target lengths may result in an inevitable quality drop of answering in many cases.

Thus, we perform experiments on two popular summarization datasets including CNNDM~\citep{hermann2015teaching} and NYT~\citep{durrett2016learning}. CNNDM contains news articles from the CNN and Daily Mail websites, with labelled abstractive and extractive summaries. There are 287,226 training samples, 13,368 validation samples and 11,490 test samples. NYT contains 110,540 articles with abstractive summaries. We follow its paper to split the original dataset into 100,834 training and 9,706 test examples. After tokenized by GPT-2 tokenizer, the reference summaries in CNNDM have an average length of 71 tokens with a standard deviation of 28 tokens, while the reference summaries in NYT have an average length of 104 tokens with a standard deviation of 28 tokens. The following subsections explain how to train and use different modules of our method. The details of hyper-parameters is in Appendix~\ref{Appendix:HP_setting}.

\subsubsection{Data processing and augmentation} \label{Sec:4.2.1}
We design a set of standard control prompts, including five control types: ``\texttt{more than ** tokens}'', ``\texttt{less than ** tokens}'', ``\texttt{equal to ** tokens}'', ``\texttt{between ** and ** tokens}'' and ``\texttt{none}''.
``\texttt{**}'' means the value of expected length from user intention, and ``\texttt{none}'' means no length constraints.
For each type, we randomly sample a target summary length from 50 to 150 tokens based on the general news summary length, and fill these lengths into ``\texttt{**}'' field of a randomly sampled SCP. To further simulate real user utterances with length control intention, around 20 possible augmented prompt templates are introduced for each SCP. Examples of templates are shown in Figure~\ref{Fig:2} and Appendix~\ref{Appendix:standard_p}. 
Finally, we can create augmented input data by replacing the placeholders in the augmented templates with target lengths and original articles.

\subsubsection{Training of standard prompt extractor} \label{Sec:4.2.2}
As introduced in Section~\ref{Sec:3.2}, we train two types of models, \textit{i.e.,} generative and discriminative models, to serve as a standard prompt extractor.
In particular, we fine-tune the GPT2-small model as a generative extractor and the BERT-small model as a discriminative extractor. Both two pre-trained checkpoints are obtained from huggingface~\citep{wolf2019huggingface}. We use the above augmented input data to fine-tune models. To make it clear, we use the original articles of CNNDM and NYT, and first sample a SCP for each article, and then sample an augmented prompt template from a pre-designed set. Next, we randomly assign the target length values between 50 and 150 to each article to form the finalized augmented template. Each original article associated with its augmented template serves as input data, and its corresponding SCP serves as the expected prediction, to finally train the standard prompt extractor.
\begin{table}[t]
  \centering
    \begin{tabular}{ccc}
    \toprule
    \textbf{Extractor} & \textbf{Acc.} & \textbf{Acc. Gen.} \\
    \midrule
    BERT-base-cls-2 & \textbf{99.9}  & \textbf{99.9} \\
    BERT-base-cls-3 & 99.7  & 99.8 \\
    GPT-small & 97.7  & 97.5 \\
    \bottomrule
    \end{tabular}%
    \caption{Evaluation on the accuracy and generalization of standard prompt extractors (SPEs). ``cls-2'' and ``cls-3'' refer to only predicting the minimum and maximum values, or predicting the control type as well. ``Acc.'' is the prediction accuracy on an in-sample test set, while ``Acc. Gen.'' denotes the generalization performance of SPEs on unseen prompt templates.}
  \label{tab:SCP}%
\vspace{-0.5em}
\end{table}%
Results of evaluating SPEs are given in Table~\ref{tab:SCP}. 
``Acc. Gen.'' means we use 30\% of randomly sampled augmented control prompts as out-of-sample templates for evaluation, and only train the SPE models on the remaining 70\% templates.
We can see that BERT-base-cls-2 can achieve almost 100.0\% test accuracy for extracting SCPs, and it also generalizes well for out-of-sample control prompts that are not seen in training. The accuracy of GPT-small is relatively lower, for which the reason may be that fully matching the whole generated texts is harder than extracting key values. The learning curves are presented in Appendix~\ref{Appendix:stan_prompt}. Overall, a well-trained SPE does not introduce much noise or performance drop in our end-to-end implementation. 
We use BERT-base-cls-2 as the discriminative extractor in later experiments to achieve clear and accurate minimum and maximum target values.

\begin{table}[tb]
  \centering
    \begin{tabular}{cccccc}
    \toprule
          & \textbf{MU} & \textbf{EQ} & \textbf{MO} & \textbf{LE} & \textbf{BT} \\
    \midrule
    \textbf{CNNDM} & 28.7 & 43.3 & 43.6  & 2.8  & 32.9 \\
    \textbf{NYT} & 22.9 & 33.7 & 19.9 & 12.9 & 21.5 \\
    \bottomrule
    \end{tabular}%
    \caption{Averaged length control errors of comparing the actual length of reference summary to our sampled length control instructions on test set.}
  \label{tab:Baseline}%
\end{table}%

\begin{table*}[htbp]
\centering{\small
    {\begin{tabular}{ccrrrrcrrrrc}
    \toprule
    \toprule
    \multirow{2}[4]{*}{\textbf{Model}} & \multirow{2}[4]{*}{\textbf{Setting}} & \multicolumn{5}{c}{\textbf{CNNDM}}             & \multicolumn{5}{c}{\textbf{NYT}} \\
\cmidrule{3-12}          &      & \multicolumn{1}{c}{\textbf{R1}$\uparrow$} & \multicolumn{1}{c}{\textbf{R2}$\uparrow$} & \multicolumn{1}{c}{\textbf{RL}$\uparrow$} & \multicolumn{1}{c}{\textbf{B.S.}$\uparrow$} & \multicolumn{1}{c}{\textbf{Error}$\downarrow$} & \multicolumn{1}{c}{\textbf{R1}$\uparrow$} & \multicolumn{1}{c}{\textbf{R2}$\uparrow$} & \multicolumn{1}{c}{\textbf{RL}$\uparrow$} & \multicolumn{1}{c}{\textbf{B.S.}$\uparrow$} & \multicolumn{1}{c}{\textbf{Error}$\downarrow$} \\
    \midrule
    \multirow{4}[2]{*}{GPT-S} 
          & Prompt & \underline{37.76}  & 15.58  & \underline{38.05}  & \underline{62.32}  & 18.16  & 47.22  & 29.47  & 42.01  & 67.76  & 17.62  \\
          & Prompt+RL & 37.52  & 15.31  & \textbf{38.79}  & \textbf{62.42}  & 14.29  & 47.30  & 29.84  & \underline{42.36}  & \underline{67.81}  & 10.53  \\
          & Prompt+filter & \textbf{38.04}  & \textbf{16.29}  & 37.12  & 62.05  & \underline{10.57}  & \textbf{47.88}  & \textbf{30.55}  & \textbf{42.50}  & \textbf{67.87}  & \underline{8.06}  \\
          & Prompt+RL+filter & 37.48  & \underline{16.01}  & 37.20  & 61.88  & \textbf{7.06}  & \underline{47.84}  & \underline{30.43}  & 42.26  & 67.54  & \textbf{3.89}  \\
    \midrule
    \multirow{4}[2]{*}{GPT-M} 
          & Prompt & \textbf{38.85}  & 15.93  & \underline{38.48}  & \textbf{63.02}  & 21.32  & 48.34  & 30.74  & 43.64  & \underline{68.75}  & 13.17  \\
          & Prompt+RL & 38.30  & 15.89  & \textbf{39.29}  & \underline{62.90}  & \underline{6.59}  & 48.23  & 30.58  & 43.61  & 68.67  & 12.61  \\
          & Prompt+filter & \textbf{38.85}  & \textbf{17.29}  & 37.68  & 62.48  & 11.21  & \textbf{49.73}  & \textbf{32.65}  & \textbf{44.55}  & \textbf{69.00}  & \underline{6.75}  \\
          & Prompt+RL+filter & 37.83  & \underline{16.89}  & 37.20  & 61.91  & \textbf{4.98}  & \underline{49.41}  & \underline{32.18}  & \underline{44.05}  & 68.40  & \textbf{3.65}  \\
    \midrule
    \multirow{4}[2]{*}{GPT-L} 
          & Prompt & 38.27  & 16.37  & \textbf{38.92}  & \textbf{63.09}  & 6.89  & 49.41  & 32.20  & \underline{44.31}  & 69.36  & 10.64  \\
          & Prompt+RL & 38.23  & 16.42  & \underline{38.86}  & \underline{63.06}  & 6.62  & 49.35  & 32.24  & \underline{44.31}  & 69.27  & 8.52  \\
          & Prompt+filter & \textbf{38.75}  & \textbf{16.85}  & 38.23  & 62.85  & \underline{3.34}  & \textbf{50.04}  & \textbf{32.65}  & \textbf{44.35}  & \underline{69.48}  & \underline{4.82}  \\
          & Prompt+RL+filter & \underline{38.70}  & \underline{16.52}  & 38.39  & 62.98  & \textbf{3.22}  & \underline{50.01}  & \underline{32.52}  & 44.14  & \textbf{69.51}  & \textbf{4.60}  \\
    \bottomrule
    \bottomrule
    \end{tabular}}%
    \caption{Comparison of methods in multiple-type control, where we consider all the four candidate types of control instructions in Table~\ref{tab:1}. In all cases, jointly using RL and sample filtering achieve the lowest control error.}
  \label{tab:MU}}%
\end{table*}%

\subsubsection{Supervised Fine-Tuning of GPT models} \label{Sec:4.2.4}

To build the baseline summarization model with length control ability, we apply three pre-trained GPT-2 models with 124M, 355M and 774M parameters from Huggingface, denoted as GPT-S, GPT-M, GPT-L, respectively. 
We randomly split the original training dataset into four parts with approximately equal size, and each is augmented with one type of SCP. 
According to the actual text length of the reference summary, 
we then randomly sample one (for ``\texttt{less than **}'' or ``\texttt{more than **}'') or two (for ``\texttt{between ** and **}'') target lengths between 50 and 150 while ensuring that the range contains the reference summary length. For the control type of ``\texttt{equal to} **'', the target value is fixed to the actual length of reference summary.
To simulate real user utterances with control instruction, we build augmented utterances by first randomly sampling prompt templates equally distributed across four control types (given in Table~\ref{tab:control_p} in Appendix), and then replacing the placeholders by the original articles and sampled target values. 
Next, we prepend the corresponding SCP to the augmented original input (separated by ``\texttt{:}'') to formulate the model input of each example. Note that SCPs can be assumed to be known when given the user's input and high accuracy of SPEs, thus the formulation of model inputs is also applicable in the inference.
Finally, we perform supervised fine-tuning on the data to enable pre-trained GPTs to summarize texts with a length control ability. 

\subsubsection{Fine-Tuning with Reinforcement Learning} \label{Sec:4.2.5}
On top of the above supervised fine-tuned GPTs, that is baseline, we further propose to improve the accuracy of length control via reinforcement learning with the PPO method as described in Section~\ref{Sec:3.4}. In other words, the backbone PLMs in our method are these supervised fine-tuned GPTs that to some extent have already owned the ability of controlling generated text lengths.
Again, for augmenting the input articles from the original datasets, we follow the similar data processing as like supervised fine-tuning mentioned above. Except that we randomly sample target lengths between 50 and 150 (not associated with reference summary length).
We use the proposed rule-based reward model with the parsed standard control information (i.e. control type and target values). 

Exploratory experiments show that actor-critic generally works better than actor-only, thus in the main experiments we use actor-critic setting. 
We apply AdamW optimizers without learning rate schedule, while the detailed hyper-parameter setting are given in Appendix.

\subsection{Results} \label{Sec:4.3}
\subsubsection{Baseline Method}
We build the length control test set by sampling control instructions for each reference summary from the test sets of both two datasets, and all the following experiments are performed on it. Similar to RL, we randomly sample target length between 50 and 150 for each example. We define length control error as the negative reward in Table~\ref{tab:1} representing the average difference between the output length and the desired range. Then we use the actual length of reference summary to calculate length control errors as shown in Table~\ref{tab:Baseline}, which can be considered as the baseline of length control errors. ``MU'' refers to test with sampled instruction equally distributed across all control types, ``EQ'', ``MO'' ``LE'', ``BE'' refer to test with sampled instructions for control types ``Equal'', ``More'' ``Less'', ``Between'', respectively. The results depend on the length distributions of labeled summaries. 






\begin{table*}[htbp]
\centering{\footnotesize
    {\begin{tabular}{ccrrrrcrrrrc}
    \toprule
    \toprule
    \multirow{2}[4]{*}{\textbf{Model}} & \multirow{2}[4]{*}{\textbf{Setting}} & \multicolumn{5}{c}{\textbf{CNNDM}}             & \multicolumn{5}{c}{\textbf{NYT}} \\
\cmidrule{3-12}          &      & \multicolumn{1}{c}{\textbf{R1}$\uparrow$} & \multicolumn{1}{c}{\textbf{R2}$\uparrow$} & \multicolumn{1}{c}{\textbf{RL}$\uparrow$} & \multicolumn{1}{c}{\textbf{B.S.}$\uparrow$} & \multicolumn{1}{c}{\textbf{Error}$\downarrow$} & \multicolumn{1}{c}{\textbf{R1}$\uparrow$} & \multicolumn{1}{c}{\textbf{R2}$\uparrow$} & \multicolumn{1}{c}{\textbf{RL}$\uparrow$} & \multicolumn{1}{c}{\textbf{B.S.}$\uparrow$} & \multicolumn{1}{c}{\textbf{Error}$\downarrow$} \\
    \midrule
    \multirow{4}[2]{*}{Equal} & Prompt & \textbf{38.14}  & 15.71  & \textbf{38.91}  & \textbf{62.62}  & 26.13  & \textbf{47.61}  & \textbf{30.36}  & \underline{42.75}  & \textbf{67.85}  & 27.98  \\
          & Prompt+RL & 35.67  & 14.64  & \underline{38.73}  & 61.86  & 13.61  & 47.57  & \underline{30.33}  & \textbf{42.88}  & \underline{67.82}  & 18.81  \\
          & Prompt+filter & \underline{37.90}  & \textbf{16.26}  & 37.42  & 61.89  & \underline{12.47}  & \underline{47.60}  & 30.32  & 42.02  & 67.80  & \underline{17.80}  \\
          & Prompt+RL+filter & 37.56  & \underline{16.10}  & 38.15  & \underline{62.23}  & \textbf{8.35}  & 47.58  & 30.29  & 42.15  & 67.71  & \textbf{8.72}  \\
    \midrule
    \multirow{4}[2]{*}{Less} & Prompt & \textbf{37.08}  & \textbf{15.74}  & \underline{36.64}  & \textbf{61.88}  & 0.47  & 46.11  & 28.96  & 41.32  & \textbf{67.07}  & 10.33  \\
          & Prompt+RL & \underline{37.03}  & 15.64  & \textbf{36.87}  & \underline{61.75}  & 0.38  & 45.75  & 28.91  & 41.08  & 66.84  & 0.96  \\
          & Prompt+filter & 36.92  & \underline{15.72}  & 35.90 & 61.17  & \underline{0.22}  & \textbf{46.68}  & \underline{29.87}  & \underline{41.53}  & 66.87  & \underline{2.09}  \\
          & Prompt+RL+filter & 36.90  & \underline{15.72}  & 35.87  & 61.13  & \textbf{0.21}  & \underline{46.65}  & \textbf{30.43}  & \textbf{42.03}  & 65.96  & \textbf{0.32}  \\
    \midrule
    \multirow{4}[2]{*}{More} & Prompt & \underline{38.00}  & 15.43  & 37.82  & \textbf{62.41}  & 39.94  & 44.01  & 27.12  & 40.22  & 66.62  & 2.27  \\
          & Prompt+RL & 35.75  & 14.83  & \textbf{38.88}  & 61.79  & \underline{13.77}  & 42.45  & 25.94  & 39.89  & 65.85  & \underline{1.32}  \\
          & Prompt+filter & \textbf{38.53}  & \textbf{16.44}  & 37.64  & 62.13  & 23.05  & \textbf{47.78}  & \textbf{30.63}  & \textbf{42.39}  & \textbf{68.00}  & 1.42  \\
          & Prompt+RL+filter & 37.43  & \underline{16.26}  & \underline{37.92}  & \underline{62.22}  & \textbf{6.01}  & \underline{47.75}  & \underline{30.53}  & \underline{42.27}  & \underline{68.94}  & \textbf{1.01}  \\
    \midrule
    \multirow{4}[2]{*}{Between} & Prompt & 36.38  & 15.03  & \underline{38.65}  & 61.96  & 5.76  & \textbf{47.65}  & \textbf{30.07}  & 41.90  & \underline{67.52}  & 18.63  \\
          & Prompt+RL & 36.10  & 14.95  & \textbf{38.99}  & 61.80  & 4.53  & 47.09  & 29.74  & \textbf{42.18}  & \textbf{67.63}  & 10.75  \\
          & Prompt+filter & \textbf{38.06}  & \textbf{16.43}  & 37.44  & \textbf{62.07}  & \underline{1.15}  & 47.13  & 29.70  & 41.37  & 67.47  & \underline{6.76}  \\
          & Prompt+RL+filter & \underline{37.85}  & \underline{16.28}  & 37.45  & \underline{62.00}  & \textbf{1.09}  & \underline{47.58}  & \underline{30.02}  & \underline{42.05}  & 67.50  & \textbf{3.18}  \\
    \bottomrule
    \bottomrule
    \end{tabular}%
    \caption{Comparison of four control types in the multiple type control setting using GPT-S on NYT datasets.}
  \label{tab:MT-S}}}%
\end{table*}

\subsubsection{Main Results} \label{Sec:4.3.1}
As Table~\ref{tab:MU} shows, we compare models with four different settings for prompt-based length control, including (1) \texttt{\textbf{Prompt}}: use GPTs with prompt-based SFT to control the output length; (2) \texttt{\textbf{Prompt+RL}}: the GPTs used in (1) but further enhanced with reinforcement learning; (3) \texttt{\textbf{Prompt+filter}}: the GPTs in (1) but equipped with sample filtering; and (4) \texttt{\textbf{Prompt+RL+filter}}: the enhanced GPTs with both RL and sample filtering, which is a combination of (2) and (3). 
For evaluation, we apply relevance scores including F1 of Rouge Scores \citep{rouge2004package} (denoted as ``R1'', ``R2'', ``RL'') and BertScore \citep{zhang2019bertscore} (denoted as ``B.S''), and length control error (denoted as ``Error''). We select the checkpoint with the lowest validation control error and less than 1 point's drop of BertScore for evaluation on the test set. For all methods with sample filtering, we set the number of output sequences to 8, and select the one with the highest reward. 

Averaged results of multi-type control are presented in Table~\ref{tab:MU}. Note that Rouge and BertScore can be less than the general state-of-the-art summarization models without length control, since our sampled length distribution can be different from reference summaries. 
In fact, the mean and standard deviation of the reference lengths are 71 and 28 tokens respectively for CNNDM, 104 and 35 tokens for NYT. The difference of control errors for two datasets can partly be due to their original length distributions. Overall, we can see that for all settings, our proposed RL method can achieve an improvement of length control ability with lower control errors. By further using sample filtering supported by the rule-based reward model, both \texttt{Prompt+filter} and \texttt{Prompt+RL+filter} can achieve lower control errors than not using sample filtering like the method (1) and (2). After checking the learning curves (see Appendix~\ref{Appendix:learning_curves}), we also find that the relevance metric BertScore indeed does not have a clear decrease trend in early stage as the validation reward increases. It indicates that with our method, the relevance of texts can be preserved as the control errors reduces during the RL fine-tuning.

\subsubsection{Comparing of different control types} \label{Sec:4.3.2}

We deconstruct the multiple-type controls and thus evaluate the effect of our proposed method on each particular control type. Results on both CNNDM and NYT are given in Table~\ref{tab:MT-S}. 
In general, our proposed methods bring a significant improvement of length control accuracy (\textit{i.e.,} Error) for all the four control types. 
Moreover, some insightful findings can be obtained from Table~\ref{tab:MT-S}. As the average length of reference summary in CNNDM (71 tokens) is much less than the average of sampled target lengths, \textit{i.e.,} 100 tokens, therefore, to generate with ``\texttt{more than}'' a sampled target length is harder than ``\texttt{less than}'' for all candidate methods. 
However, the \texttt{Prompt+RL+filter} can still provide a significantly large improvement on the control type of ``\texttt{more than}'', by reducing the Error from $41.9$ to $6.0$. In the case of ``\texttt{less than}'' with sample filtering, the RL method does not further reduce the validation error as it is already quite low, thus the default checkpoint is always selected even after RL fine-tuning.

\subsection{Generalization to unseen templates} \label{Sec: 4.3}

To evaluate if the tuned model can generalize to unseen prompt templates of length control, we conduct an extra experiment by tuning on a 70\% subset of prompt templates randomly sampled from Table~\ref{tab:control_p} in the Appendix, and check our model performance with the rest test of unseen prompt templates, as give in Table~\ref{tab:gen_prompt}. The difference between ``In-sample'' and ``Out-sample'' setting is whether the out-of-sample set of control prompt templates is applied for training. We notice that in some cases, there is a slight performance degradation on out-of-sample prompt templates, but the length control ability is still significantly better than baseline method. This demonstrates that our proposed method has strong generalization to novel prompt templates. We believe with a larger set of prompt templates in training, this generalization power can still be largely improved.
\begin{table}[tbp]
  \centering{\footnotesize
    \begin{tabular}{p{3em}p{4.8em}p{1.2em}p{1.2em}p{1.2em}p{1.2em}p{1.6em}}
    \toprule
    \toprule
    \textbf{Type}  & \textbf{Setting} & \textbf{R1}    & \textbf{R2}    & \textbf{RL}    & \textbf{B.S.}  & \textbf{Error}$\downarrow$ \\
    \midrule 
    \multirow{3}[2]{*}{NYT} & Baseline & 47.2 & 29.5 & 42.0 & 67.8 & 17.6  \\
          & In-sample & 47.8 & 30.4 & 42.3 & 67.5 & 3.9 \\
          & Out-sample & 47.7  & 30.2  & 42.3  & 67.1  & 4.1  \\
    \midrule
    \multirow{3}[2]{*}{CNNDM} & Baseline & 37.8  & 15.6  & 38.1  & 62.3  & 14.7  \\
          & In-sample & 37.6  & 15.3  & 38.8  & 62.3  & 7.6  \\
          & Out-sample & 37.7  & 15.4  & 38.7  & 62.4  & 8.1  \\
    \bottomrule
    \bottomrule
    \end{tabular}%
      \caption{Generalization to out-of-sample control templates of GPT-S for multi-type length control.}
  \label{tab:gen_prompt}}%
\end{table}%

\section{Discussion} \label{Sec:6}

\subsection{Quality of the generated summaries}
We have checked the generated summaries under length control in the log file, where we printed out a subset of generated summaries in each epoch of the validation stage and the test stage. We confirm that the summaries generated by the final model are coherent summaries without any meaningless repetition or sudden cut-offs. In fact, our sample filtering method does not update the parameters of GPTs, thus the informativeness, perplexity, coherence are preserved. Our RL-based tuning method updates parameters through log-probabilities given by the output layer, while the parameters work on the embedding dimension, which are shared across all tokens. In this case, the n-gram rouge scores are strong indicators of the perplexity change. Thus, a little change of Rouge/Bertscore will not cause a significant change of coherence. In addition, we can add a SFT loss to avoid quality decrease, and the experiments are given in the Appendix ~\ref{Appendix:SFT_Loss}. 

\subsection{Performance with larger models}
We believe our method will still work well for larger pre-trained language models. This is because larger models are more powerful in learning length control abilities given the accurate feedbacks. Additionally, we can develop a much larger tuning dataset to do RL for more accurate control. This is mostly an engineering work. To make the pretrained model sensitive to the length control instructions, some first-stage prompt-based tuning may be still needed. However, this requires much higher computational power. As we know, in many cases, small models like GPT-2, Flan T5 \citep{Chung2024Scaling}, Tiny-Llama \citep{Zhang2024TinyLlama} also works well in tasks like summarization. If we only need to do summarization in applications like news-reading software, 0.5B-1B model like GPT-2 or compressed LLMs can be sufficient.

\section{Conclusion} \label{Sec:7}
We proposes a method for improving the length control ability of GPT-style PLMs under multiple control types, especially for the domain of text summarization.
The standard prompt extractor and rule-based reward model are introduced to provide an accurate control signal for both fine-tuning and inference. We apply a modified PPO algorithm for enhancing the length controlled generation. In the inference, sample filtering is further introduced for selecting a generated sample that follows the instruction. The method is proved to be effective for three sizes of GPT-2 models on both CNNDM and NYT summarization datasets. Compared to the baseline using prompt-based strategies on GPTs, our method further achieves a significant improvement in terms of control accuracy. Moreover, it can process diverse length-control prompts with strong generalization ability to new prompt templates, and can naturally adapt to most LLMs for improving user experience.

\section{Limitations} \label{Sec:8}
The limitations of our study involve the following aspects. First, similar to RLHF implemented in InstructGPT, finetuning with RL may result in a decrease of the language modeling evaluation metric. Well designed in-context learning or introducing adaptors/LoRA particularly tuned for length control may be potential solutions for this. Second, the control performance relies on the goodness of standard prompt extractor. When the generative one is applied, it is possible to generate outputs that can not be fully parsed with rule-based method. Third, when the discriminator is applied for filtering the generated samples in inference, usually a large beam size is required, thus longer inference time and computing cost may needed. As the probability distribution across all tokens are available in auto-regressive generation, this extra cost can be well scaled. 


\newpage

\bibliography{Reference}
\newpage

\clearpage

\appendix
\newpage
\section{Appendix} \label{Appendix}

\subsection{Plots of control errors for all the four control types.}

\begin{figure}[tbh]
\begin{center}
 \includegraphics[width=0.8\linewidth]{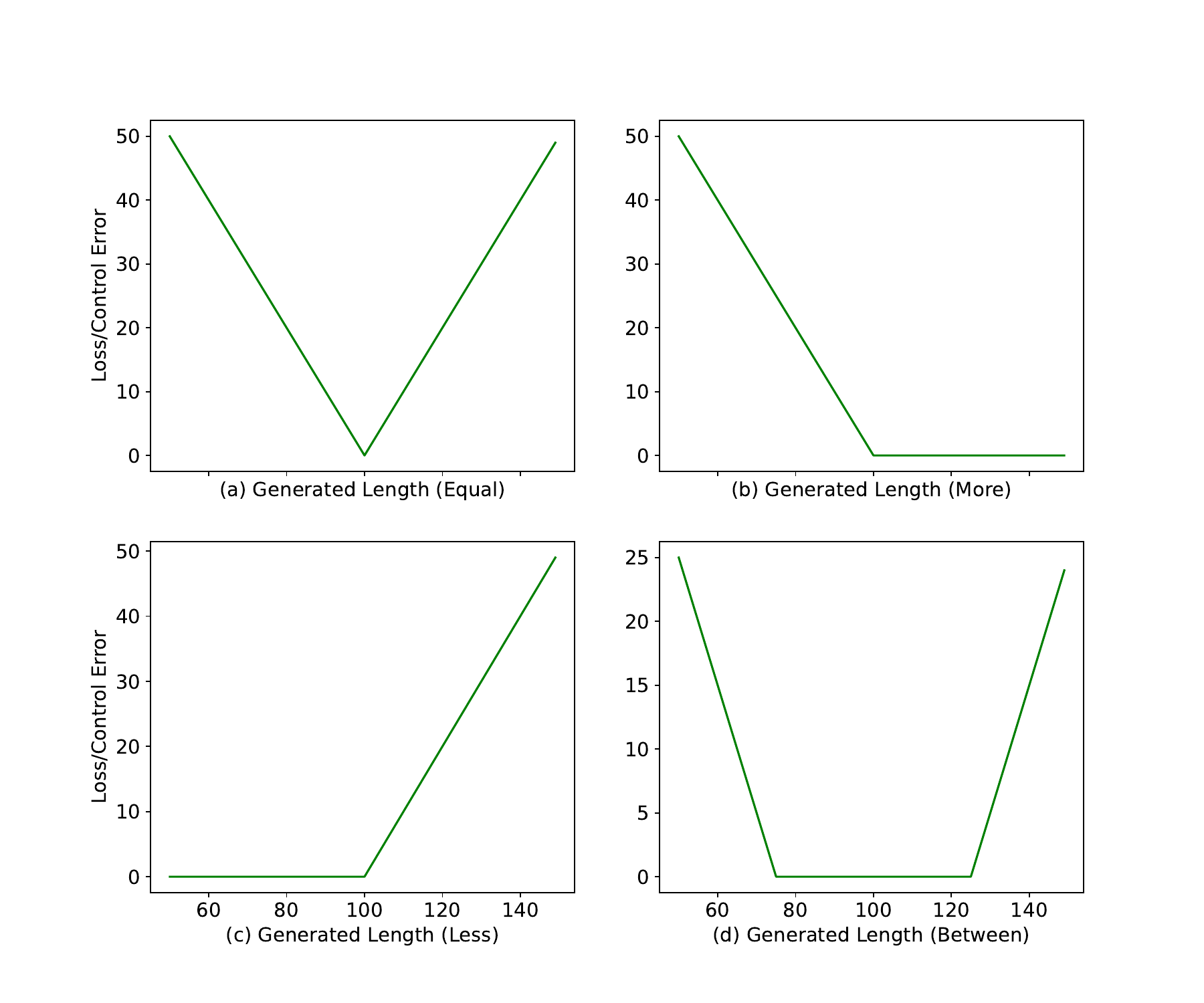}
 \caption{Plots of control error functions, which is the negative of reward functions. } \label{Fig:control_error}
\end{center}
\end{figure}
To better illustrate the reward functions shown in Table~\ref{tab:1}, we provide the plots of control error functions in Figure~\ref{Fig:control_error}. We set the target length $L_t = 100$ for the case of ``Equal'', ``More'' and ``Less'', and set the upper bound and lower bound $L_L=75$ and $L_U=125$ for the case of ``Between''. We change the length of generated sequence ``L\_g'' from 50 to 150 and show the corresponding control error in each case using the connected curves. We can see that in the ranges that satisfy the control requirement, no error or negative reward occurs. Thus the parameters are not updated based on the corresponding examples. As the deviance becomes larger, the loss will also be larger.

\subsection{Algorithm for length controlled fine-tuning with our modified PPO}\label{Appendix:Alg}
Following the explanations in Section~\ref{Sec:3.4}, we further provide an algorithm table for our modified PPO fine-tuning in Algorithm~\ref{Alg:1}. Note that this algorithm does not include the training of SPEs and sample filtering. In practice, we stop the tuning process of PPO when the validation BERTscore drop for more than 1.0 point.
\begin{algorithm*}[htbp]
\caption{Algorithm for controlled fine-tuning with modified PPO}
\begin{algorithmic}[1]
  \STATE Get a pre-trained GPT model to initialize the policy network $\pi_{\theta_{old}}(a|s)$.
  \STATE Initialize critic network $Q_{\theta}(s', a)$.
  \STATE Initialize hyper-paramaters $N_{iteration}$, $M$, $B$, $n_{epoch}$, $c$, $\beta$.
  \FOR{i<=1,...,$N_{iteration}$}
   \FOR{j=1,...,M}
        \STATE Get an input sequence $s_0$ augmented with random sampled augmented control prompt from the data-loader.
        \STATE Run SPE to get the SCP $s'$ from the input sequence.
        \STATE Run policy $\pi_{\theta_{old}}(a|s)$ for an input sequence with augmented control prompt $s$ to get an output sequence $a$, policy $\pi_{\theta_{old}}$.
        \STATE Get the reward of output sequence $a$ with reward model $r=r(s',a)$.
        \STATE Store input $s$, SCP $s'$, generate sequence $a$, reward $r$ and old policy $\pi_{\theta_{old}}$ into memory.
    \ENDFOR
   \FOR{e=1,...,$n_{epoch}$}
       \FOR{b=1,...,B}
           \STATE Take the $b$-th mini-batch $(s', a, r, \pi_{\theta_{old}})$ from the memory.
           \STATE Use the actor and critic networks to get the new policy and value $\pi_{\theta}(a|s), Q_{\theta}(s', a)$.
           \STATE Compute the ratio $r(\theta)=\frac{\pi_{\theta}(a|s)}{\pi_{\theta_{old}}(a|s)}$.
           \STATE Compute advantage estimate $\hat{A} = r-Q_{\theta_{old}}(s', a)$.
           \STATE Compute $L^{CLIP}$ with Eq.~\eqref{Eq:2}.
           \STATE Compute the KL-divergence $D_{KL}(\pi_{\theta}|\pi_{\theta_{old}})$. 
           \STATE Compute the Entropy $S[\pi_{\theta}|(s)]$.
           \STATE Compute the actor loss $L_{\theta}^{A}$ with Eq.~\eqref{Eq:3}.
           \STATE Update the policy network parameters $\theta$ with gradients of $L_{\theta}^{A}$.
           \STATE Compute the value loss $L^Q_{\theta} = MSE(Q_{\theta}(s', a), r)$.
           \STATE Update the critic network parameters $\theta$ with gradients of $L^V_{\theta}$.
        \ENDFOR
    \ENDFOR
  \ENDFOR
  \RETURN $\theta$
\end{algorithmic}
\label{Alg:1}
\end{algorithm*}


\subsection{Examples of standard control prompt and augmented control prompt templates} \label{Appendix:standard_p}

The SCPs and corresponding augmented prompt templates for generating the augmented input with length control information are given in Table~\ref{tab:control_p}. In the experiments, we use the augmented prompts to train and evaluate the standard prompt extractor. For the backbone PLMs and reward models, SCPs can be considered as available, given the high performance of SCPs. 
\begin{table*}[htp]
  \centering{\scriptsize
    \begin{tabular}{p{12em}p{12em}p{12em}p{12em}}
    \toprule
    \multicolumn{1}{c}{\textbf{Equal}} & \multicolumn{1}{c}{\textbf{Less}} & \multicolumn{1}{c}{\textbf{More}} & \multicolumn{1}{c}{\textbf{Between}} \\
    \midrule
    summarize "*" with length ? & summarize "*" with length smaller than ? & summarize "*" with length larger than ! & summarize "*" with length between ! and ? \\
    summarize the following document with length ?: "*" ' & summarize the following document with length smaller than !: "*" & summarize the following document with length larger than !: "*"  & summarize the following document with length between ! and ?: "*" \\
    Summarize with exactly ? tokens: *' & Summarize with less than ? tokens: * & Summarize with more than ! tokens: * & Summarize with between ! and ? tokens: * \\
    I want a summary of "*" with exactly ? Tokens & I want a summary of "*" with less than ? Tokens & I want a summary of "*" with more than ! Tokens 
    & I want a summary of "*" with between ! and ? Tokens \\
    Give me a summary with ? tokens from "*"' & Give me a summary with less than ? tokens from "*" & Give me a summary with more than ! tokens from "*" & Give me a summary with between ! and ? tokens from "*" \\
    Please summarize "*" with exactly ? Tokens &Please summarize "*" with less than ? Tokens & Please summarize "*" with more than ! Tokens & Please summarize "*" with between ! and ? Tokens \\
    Write a summary of "*" with exactly ? Tokens & Write a summary of "*" with less than ? Tokens & Write a summary of "*" with more than ! Tokens & Write a summary of "*" with between ! and ? Tokens \\
    summarize "*" with ? tokens for me & summarize "*" with less than ? tokens for me & summarize "*" with more than ! tokens for me & summarize "*" with between ! and ? tokens for me \\
    Please give me a summary of "*" with ? Tokens & Please give me a summary of "*" with less than ? Tokens & Please give me a summary of "*" with more than ! Tokens & Please give me a summary of "*" with between ! and ? Tokens \\
    I need a summary of length ? for "*" & I need a summary of length smaller than ? for "*" & I need a summary of length greater than ! for "*" & I need a summary of length between ! and ? for "*" \\
    generate a summary for "*" with length ? & I need a summary of length less than ? for "*" & I need a summary of length larger than ! for "*" & Need a summary of "*" with length between ! and ? \\
    Need a summary of "*" with length equal to ? & Need a summary of "*" with length smaller than ? & Need a summary of "*" with length larger than ! & write a summary of length between ! and ? for "*" \\
    write a summary of length ? for "*" & summarize the following article with no longer than ? tokens: "*" & summarize the following article with longer than ! tokens: "*" & summarize with length between ! and ?: "*" \\
    summarize with length equal to ?: "*"' & summarize the following article with shorter than ? tokens: "*" & write a summary of length larger than ! for "*" & summarize with between ! and ? tokens:"*" \\
   summarize with exactly ? tokens:"*" & write a summary of length smaller than ? for "*" &summarize with length larger than !: "*" & summarize with ! to ? tokens:"*" \\
    summarize this document with about ? tokens: "*"      & summarize with length smaller than ?: "*" & summarize with more than ! tokens:"*" & summarize "*" with ! to ? Tokens \\
    summarize "*" with around ? tokens       & summarize with less than ? tokens:"*" &  summarize the following article with over ? tokens:"*"     & Please summarize "*" with ! to ? Tokens \\
    need a summary of "*" with length ?      & summarize "*" within ? tokens  &  summarize "*" with over ? tokens & summarize following article with ! to ? tokens: "*" \\
    \bottomrule
    \end{tabular}%
    \caption{Examples of standard control prompts and corresponding augmented prompt templates, where each column shows one type of SCP followed by augmented prompt templates. Where ``*'' is the placeholder for the input article to be summarized, ``!'' and ``?'' are the placeholders for the sampled length values. To build the input examples in training and evaluation datasets, we only need to first replace ``!'' and ``?'' with the minimum and maximum target lengths, and then replace ``*'' with the original article to be summarized. }
  \label{tab:control_p}}%
\end{table*}%

\subsection{Hyper-parameter settings} \label{Appendix:HP_setting}

In this section, we provide hyper-parameter settings of different modules and training stages of our method, where we denote hyper-parameter as ``HP'' in the tables. For the standard prompt extractor, the hyper-parameter settings are given in Table~\ref{tab:HP_1}. For the trainable reward models, the hyper-parameter settings are given in Table~\ref{tab:HP_2}. For pretraining of GPT summarization models with control prompts, the hyper-parameter settings are given in Table~\ref{tab:HP_3}. For enhancing control ability with reinforcement finetuning, the hyper-parameter setting are given in Table~\ref{tab:HP_4}. 
\begin{table}[H]
  \centering{\footnotesize{
    \begin{tabular}{ccc}
    \toprule
    \textbf{HP} & \textbf{BERT extractor} & \textbf{GPT extractor} \\
    \midrule
    pretrained model & BERT-small & GPT-small \\
    optimizer & AdamW & AdamW \\
    batch size & 32    & 64 \\
    lr    & 2E-05 & 2E-05 \\
    $\beta_1$ & 0.9   & 0.9 \\
    $\beta_2$ & 0.999 & 0.999 \\
    weight decay & 1E-07 & 0 \\
    num iterations & 200k    & 200k \\
    \bottomrule
    \end{tabular}%
    \caption{Hyper-parameter setting of Standard Prompt Extractors.}
  \label{tab:HP_1}}}%
\end{table}%
\begin{table}[H]
  \centering{\footnotesize{
    \begin{tabular}{ccc}
    \toprule
    \textbf{HP} & \textbf{BERT reward} & \textbf{GPT reward} \\
    \midrule
    pretrained model & BERT-large & GPT-medium \\
    optimizer & AdamW & AdamW \\
    batch size & 64    & 32 \\
    lr    & 0.00005 & 0.00005 \\
    $\beta_1$ & 0.9   & 0.9 \\
    $\beta_2$ & 0.999 & 0.999 \\
    weight decay & 0     & 0 \\
    num iterations & 200k    & 200k \\
    \bottomrule
    \end{tabular}%
    \caption{Hyper-parameter setting of trainable reward models.}
  \label{tab:HP_2}}}%
\end{table}%
\begin{table}[H]
  \centering{\footnotesize{
    \begin{tabular}{cccc} 
    \toprule
    \textbf{HP} & \textbf{GPT-S} & \textbf{GPT-M} & \textbf{GPT-L} \\
    \midrule
    optimizer & AdamW & AdamW & AdamW \\
    batch size & 64    & 64    & 64 \\
    lr    & 5E-05 & 5E-05 & 2E-05 \\
    $\beta_1$ & 0.9   & 0.9   & 0.9 \\
    $\beta_2$ & 0.999 & 0.999 & 0.999 \\
    weight decay & 1E-06 & 1E-06 & 1E-06 \\
    num iterations & 200k  & 200k  & 200k \\
    \bottomrule
    \end{tabular}%
    \caption{Hyper-parameter setting of prompt-based SFT on pretrained GPT models.}
  \label{tab:HP_3}}}%
\end{table}%

\begin{table}[H]
  \centering{\footnotesize{
    \begin{tabular}{cccc} 
    \toprule
    \textbf{HP} & \textbf{GPT-S} & \textbf{GPT-M} & \textbf{GPT-L} \\
    \midrule
    optimizer & AdamW & AdamW & AdamW \\
    actor\_lr & 3E-07 & 3E-07 & 3E-07 \\
    critic\_lr & 0.0003 & 0.0003 & 0.0003 \\
    $\beta_1$ & 0.9   & 0.9   & 0.9 \\
    $\beta_2$ & 0.999 & 0.999 & 0.999 \\
    actor\_adam\_eps & 1E-07 & 1E-07 & 1E-07 \\
    critic\_adam\_eps & 1E-07 & 1E-07 & 1E-07 \\
    weight decay & 0     & 0     & 0 \\
    epochs & 1     & 1     & 1 \\
    update timestep & 512   & 512   & 512 \\
    surrogate epoch & 16    & 16    & 16 \\
    surrogate batch size & 32    & 16    & 8 \\
    $\beta$ & 0.1   & 0.1   & 0.1 \\
    $c$  & 0.01   & 0.01   & 0.01 \\
    $\epsilon_{clip}$ & 0.2   & 0.2   & 0.2 \\
    $\lambda$  & 1.0   & 1.0   & 1.0 \\
    \bottomrule
    \end{tabular}%
    \caption{Hyper-parameter setting reinforcement learning for pretrained GPT models. $\epsilon_clip$ is the clipping parameter $\epsilon$ shown in Eq.~\eqref{Eq:2}. $\beta$ and $c$ are weights for KL divergence and entropy as shown in Eq.~\eqref{Eq:3}. $\lambda$ is the coefficient for SFT loss.}
  \label{tab:HP_4}}}%
\end{table}%

\subsection{Extra Results} \label{Appendix:extra_results}

\subsubsection{Single-type control}

We also conduct experiments for traditional \textbf{single-type control}, where we only consider the strict SCP of ``\texttt{equal to}'' in both SFT and reinforcement fine-tuning. In details, for each example we randomly sample a augmented control prompt under the type of ``equal'' and replace the text placeholder with the input text and replace the length placeholder with the real text length of reference summary. Finally, we prepend the SCP before the main context of the augmented input. The results are given in Table~\ref{tab:SG}. Again, we can see that for all settings, the proposed RL method can provide an improvement of length control ability with lower control errors. By further using sample filtering supported by the rule-based reward model, both the basic prompt-based length control model \texttt{Prompt+filter} and the one with RL enhancement \texttt{Prompt+RL+filter} can achieve lower control errors than not using sample filtering. This demonstrate the effectiveness of both RL-based finetuning and sample filtering in this relatively simple case.

\subsubsection{Effect of SFT loss} \label{Appendix:SFT_Loss}

As was discussed in Section ~\ref{Sec:3.4}, the actor loss involves a term of SFT loss, which is controlled by $\lambda$. We conduct an extra experiment on CNNDM by comparing the tuned GPT-S models using different $\lambda$s for both the case of single and multiple control types. The results are given in Table~\ref{tab:lambda}, which shows that a suitable $\lambda$ is helpful in perserving the performance on downstream task, and the control accuracy will not be largely affected in most cases. Also, the optimal value of $\lambda$ differs in the cases of SG and MU, thus hyper-parameter tuning is usually needed.

\subsubsection{Comparing between actor-critic model and actor only model} \label{Appendix:actor_critic_ext}

Another experiment is done to check the effect of using actor-critic model in comparison with actor-only model. The details of these two settings has been discussed in Section~\ref{Sec:2.1}. We conduct experiments with both settings, and consider fine-tuning GPT-small model for single-type control. The results on NYT amnd CNNDM are given in Table~\ref{tab:actor_critic_ext_1} and Table~\ref{tab:actor_critic_ext_2}, respectively. For the case without sample filtering, the model trained with actor-critic RL perform better than the model trained with actor-only RL in terms of control accuracy on both datasets. With sample filtering, actor-critic method still significantly outperforms actor-only method on NYT, but slightly worse than actor-only method on CNNDM. On NYT, rule-based reward model achieves the lowest and second lowest in the cases with and without sample filtering respectively. Meanwhile, the trainable reward models also works well. 

\begin{table}[tbp]
\centering{\footnotesize{
    \begin{tabular}
    {p{8em}p{1.2em}p{1.2em}p{1.2em}p{1.2em}p{1.6em}}
    \toprule
    \toprule
   \textbf{Settings} & \textbf{R1}$\uparrow$    & \textbf{R2}$\uparrow$    & \textbf{RL}$\uparrow$    & \textbf{B.S.}$\uparrow$  & \textbf{Error}$\downarrow$  \\
    \midrule
    Prompt & 47.4  & 29.2  & 42.3  & 67.7  & 13.5  \\
    +RL+Rule (A-C) & 47.7  & 29.5  & 42.7  & 67.9  & 12.8 \\
    +RL+Rule (A) & 47.6  & 29.5  & 42.0  & 67.9  & 12.9 \\
    +Filter & 48.4  & 30.8  & 42.7  & 67.9  & 10.3 \\
    +RL+Filter (A-C) & 48.3  & 30.9  & 42.8  & 67.9  & \textbf{9.6} \\
    +RL+Filter (A) & 47.8  & 30.1  & 42.1  & 67.6  & \underline{9.7}  \\
    \bottomrule
    \bottomrule
    \end{tabular}}%
    \caption{The comparison of control performance of GPT-S for single-type control (``equal to'') after fine-tuning by RL w/o critic models (NYT).}
  \label{tab:actor_critic_ext_1}}%
\end{table}%
\begin{figure*}[tbh]
\begin{center}
 \includegraphics[width=0.8\linewidth]{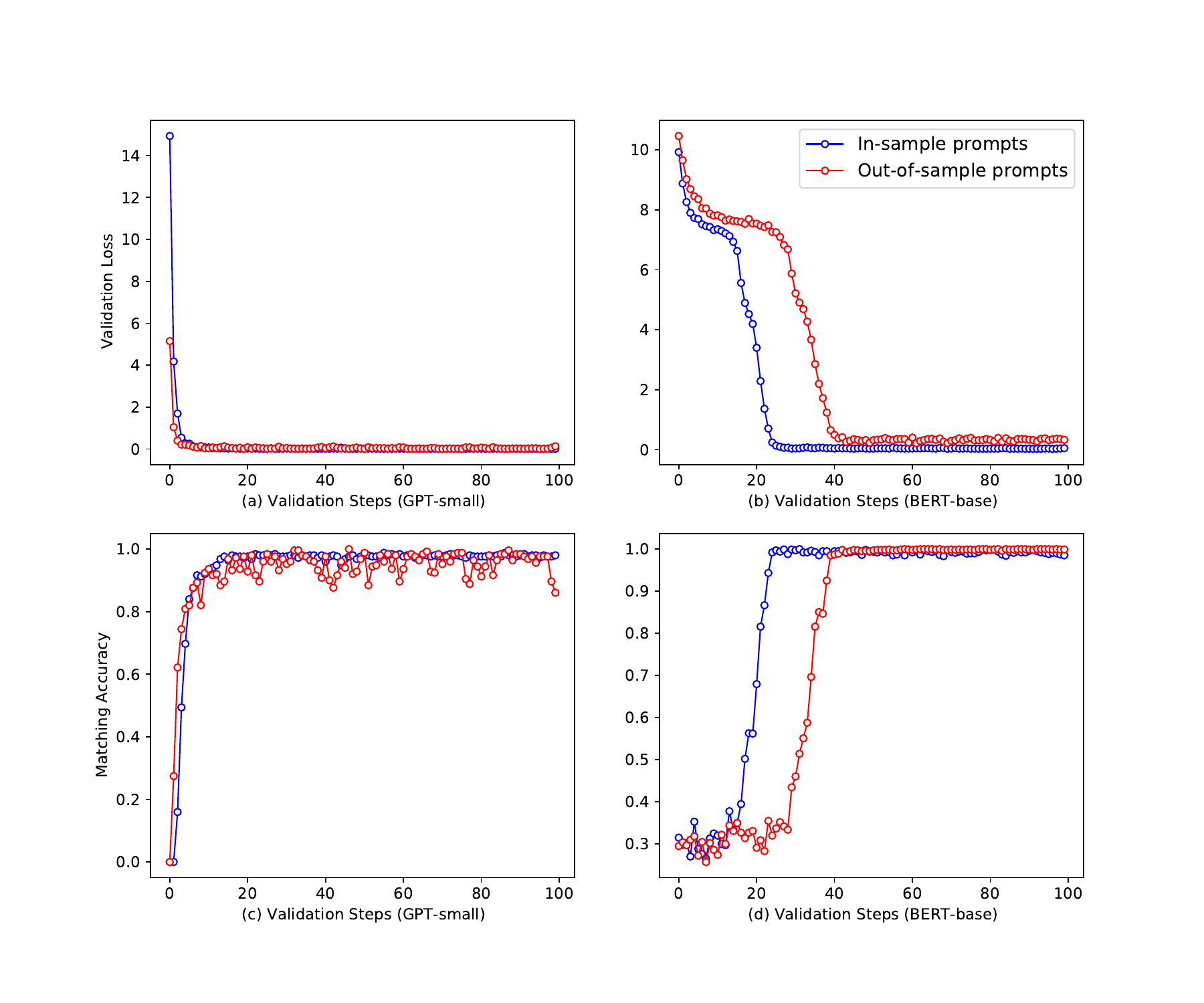}
 \caption{Learning Curves of Standard Prompt Extractors. (a) Validation losses of GPT extractor. (b) Validation losses of BERT extractor. (c) Matching accuracy of GPT extractor. (c)  Matching accuracy of BERT extractor. We show the curves of validation cross entropy and matching rate for both cases. } \label{Fig:curves_std_ext}
\end{center}
\end{figure*}

\begin{table}[tbp]
\centering{\footnotesize{
    \begin{tabular}
    {p{8em}p{1.2em}p{1.2em}p{1.2em}p{1.2em}p{1.6em}}
    \toprule
    \toprule
   \textbf{Settings} & \textbf{R1}$\uparrow$    & \textbf{R2}$\uparrow$    & \textbf{RL}$\uparrow$    & \textbf{B.S.}$\uparrow$  & \textbf{Error}$\downarrow$  \\
    \midrule
    Prompt & 37.6  & 15.2  & 37.6  & 62.3  & 11.9  \\
    +RL+Rule (A-C) & 37.3  & 14.9  & 38.9  & 61.8  & \underline{7.4}  \\
    +RL+Rule (A) & 37.7  & 15.6  & 38.2  & 62.3  & 11.0  \\
    +Filter & 38.26  & 16.1  & 37.4  & 61.9  & 10.5  \\
    +RL+Filter (A-C) & 37.3  & 15.7  & 38.8  & 61.2  & \textbf{6.3}  \\
    +RL+Filter (A) & 38.7  & 16.6  & 38.6  & 62.1  & 9.6  \\
    \bottomrule
    \bottomrule
    \end{tabular}}%
    \caption{The comparison of control performance of GPT-S for single-type control (``equal to'') after fine-tuning by RL w/o critic models (CNNDM).}
  \label{tab:actor_critic_ext_2}}%
\end{table}%

\begin{table*}[ht]
  \centering{\small
    \begin{tabular}{ccrrrrcrrrrc}
    \toprule
    \toprule
    \multirow{2}[4]{*}{\textbf{Model}} & \multirow{2}[4]{*}{\textbf{Setting}} & \multicolumn{5}{c}{\textbf{CNNDM}}             & \multicolumn{5}{c}{\textbf{NYT}} \\
\cmidrule{3-12}          &       & \multicolumn{1}{c}{\textbf{R1}$\uparrow$} & \multicolumn{1}{c}{\textbf{R2}$\uparrow$} & \multicolumn{1}{c}{\textbf{RL}$\uparrow$} & \multicolumn{1}{c}{\textbf{B.S.}$\uparrow$} & \multicolumn{1}{c}{\textbf{Error}$\downarrow$} & \multicolumn{1}{c}{\textbf{R1}$\uparrow$} & \multicolumn{1}{c}{\textbf{R2}$\uparrow$} & \multicolumn{1}{c}{\textbf{RL}$\uparrow$} & \multicolumn{1}{c}{\textbf{B.S.}$\uparrow$} & \multicolumn{1}{c}{\textbf{Error}$\downarrow$} \\
    \midrule
    \multirow{4}[2]{*}{GPT-S}
          & Prompt & 37.57  & 15.30  & 37.74  & 62.47  & 11.62  & 47.48  & 29.27  & 42.36  & 67.86  & 13.33  \\
          & Prompt+RL & 37.44  & 15.02  & 39.05  & 62.10  & \underline{7.81}  & 47.59  & 29.41  & 42.66  & 67.82  & 11.92  \\
          & Prompt+filter & 38.20  & 16.02  & 37.31  & 61.96  & 10.44  & 48.37  & 30.83  & 42.72  & 67.96  & \underline{10.30}  \\
          & Prompt+RL+filter & 37.56  & 15.85  & 38.47  & 61.53  & \textbf{6.22}  & 48.31  & 30.94  & 42.82  & 67.98  & \textbf{9.55}  \\
    \midrule
    \multirow{4}[2]{*}{GPT-M}
          & Prompt & 38.05  & 16.15  & 37.81  & 62.93  & 14.31  & 48.34  & 30.53  & 43.11  & 68.54  & 5.12  \\
          & Prompt+RL & 37.73  & 15.98  & 38.07  & 62.62  & \underline{11.57}  & 48.86  & 31.19  & 43.98  & 69.09  & 4.47  \\
          & Prompt+filter & 38.18  & 16.55  & 37.14  & 62.32  & 12.60  & 48.53  & 30.95  & 43.33  & 68.55  & \underline{2.12}  \\
          & Prompt+RL+filter & 37.91  & 16.33  & 36.97  & 62.23  & \textbf{11.33}  & 48.76  & 31.09  & 43.38  & 68.80  & \textbf{1.60}  \\
    \midrule
    \multirow{4}[2]{*}{GPT-L}
          & Prompt & 40.27  & 17.33  & 39.67  & 63.96  & 12.20  & 49.98  & 32.43  & 44.65  & 69.44  & 5.89  \\
          & Prompt+RL & 39.49  & 16.42  & 39.02  & 63.38  & \underline{9.84}  & 49.12  & 30.86  & 43.59  & 69.03  & \underline{5.54}  \\
          & Prompt+filter & 39.52  & 17.33  & 38.64  & 63.22  & 11.57  & 47.22  & 31.77  & 43.29  & 69.02  & 5.76  \\
          & Prompt+RL+filter & 39.75  & 17.18  & 38.60  & 63.15  & \textbf{8.96}  & 49.82  & 31.68  & 42.48  & 68.72  & \textbf{3.29}  \\
    \bottomrule
    \bottomrule
    \end{tabular}}%
    \caption{Comparison of methods in the setting of single-type control instruction, i.e., ``\texttt{equal to}''.}
  \label{tab:SG}%
\end{table*}

\begin{table*}[htb]
  \centering{\small
    \begin{tabular}{ccccccccccc}
    \toprule
    \toprule
    \multirow{2}[4]{*}{$\lambda$} & \multicolumn{5}{c}{\textbf{SG}}  & \multicolumn{5}{c}{\textbf{MU}} \\
\cmidrule{2-11}          & \textbf{R1}    & \textbf{R2}    & \textbf{RL}    & \textbf{B.S.}  & \textbf{Error}$\downarrow$ & \textbf{R1}    & \textbf{R2}    & \textbf{RL}    & \textbf{B.S.}  & \textbf{Error}$\downarrow$ \\ \\
    \midrule
    0.01  & 36.87  & 15.17  & 37.23  & 62.10  & 8.93  & 37.28  & 15.42  & 38.55  & 62.18  & 15.16  \\
    0.03  & 36.69  & 14.83  & 37.06  & 61.89  & 8.93  & 37.81  & 15.95  & 38.94  & 62.39  & 18.04  \\
    0.1   & 37.36  & 15.20  & 37.35  & 62.29  & 8.54  & 36.85  & 15.24  & 37.99  & 61.78  & 14.38  \\
    0.3   & 37.87  & 15.52  & 37.92  & 62.44  & 7.97  & 36.54  & 15.07  & 37.76  & 61.69  & 14.55  \\
    1     & 37.92  & 15.83  & 37.57  & 62.26  & 7.78  & 37.06  & 15.26  & 38.00  & 61.92  & 14.57  \\
    3     & 38.09  & 15.96  & 37.71  & 62.29  & 7.95  & 37.09  & 15.36  & 37.78  & 61.94  & 15.16  \\
    \bottomrule
    \bottomrule
    \end{tabular}%
    \caption{The effect of SFT loss. $\lambda$ is the hyper-parameter discussed in Section~\ref{Sec:3.4}. }
  \label{tab:lambda}}%
\end{table*}%


\newpage
\subsection{Learning curves of SPEs.} \label{Appendix:stan_prompt}

For training SPEs, we fine-tune the GPT2-small model as a generative extractor and the BERT-small model as a discriminative extractor. Note that we only predict the minimum and maximum target values with BERT.
We use the original articles of CNNDM and NYT, and first sample a SCP for each article, and then sample an augmented prompt template from the pre-designed set given in Table ~\ref{tab:control_p}. Next, we randomly assign the target length values between 50 and 150 to each article to form the finalized augmented template. Each original article associated with its augmented template serves as input data, and its corresponding SCP serves as the expected prediction or the label, to finally train the standard prompt extractor.

For GPT-based extractor, the accuracy is 1 only if the generated SCP exactly matches the label. For BERT-based extractor, we calculate the validation accuracy on a case-by-case basis: If the ground truth SCP type is ``none'', the accuracy is always 1; if the ground truth SCP type is ``more than'', we only match the minimum value and check if the minimum value is smaller than maximum value; if the ground truth SCP type is ``less than'', we only match the maximum value and check if the minimum value is smaller than maximum value; if the ground truth SCP type is ``equal to'' or ``between'', we match both of minimum and maximum values. We provide the learning curves of two types of SPEs in Figure~\ref{Fig:curves_std_ext}. 
As is shown in Figure~\ref{Fig:curves_std_ext}, both of the SPEs converge well with a validation proportion of matching rate close to 100\% in later validation steps. Meanwhile, we find the both BERT and GPT-based extractors performs fairly well on out-of-sample augmented prompts, which demonstrates strong generalization ability to new control prompts. For BERT-base, the validation curve and accuracy curve of model on out-of-sample augmented prompts converge slower than in-sample augmented prompts with a right-shift, but the accuracy values in later steps can even surpass that of in-sample validation curve. Notes than we only fine-tune the pre-trained GPT-small and BERT-base from Huggingface, which indicates the noise introduced by the extractors can generally be neglected in practice with same or larger models.

\newpage
\subsection{Learning curves of Reinforcement Fine-tuning} \label{Appendix:learning_curves}

\begin{figure*}[thb]
\begin{center}
 \includegraphics[width=1.0\linewidth]{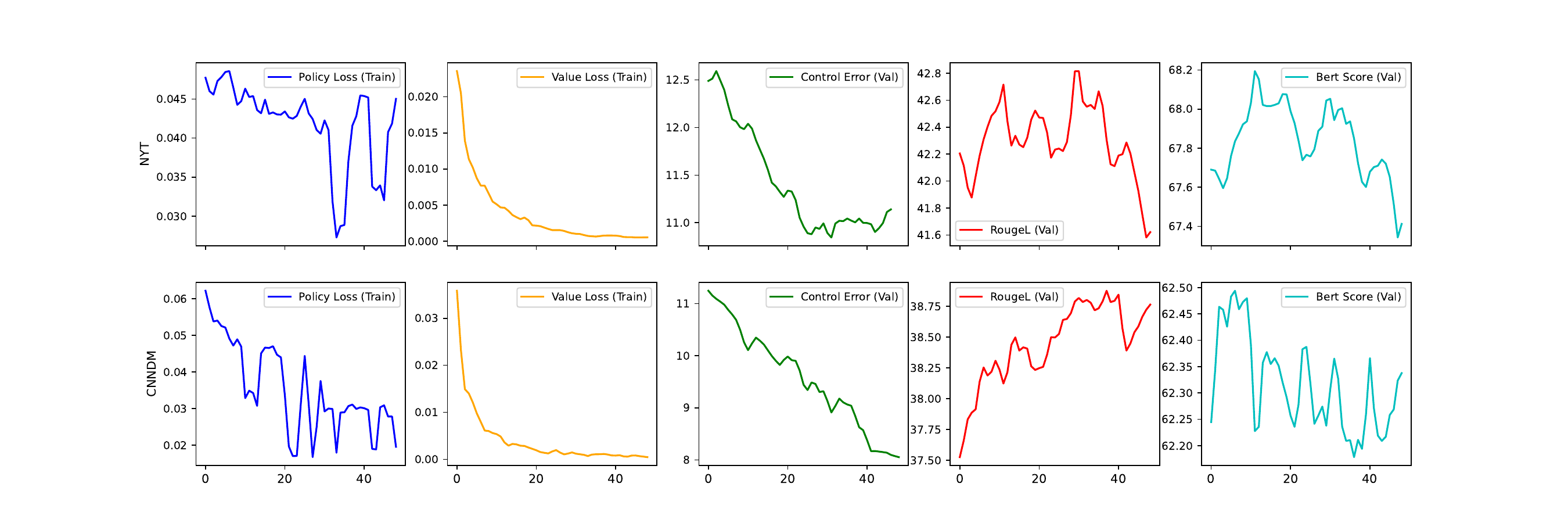}
 \caption{The Diagram of Learning Curves with GPT-S for single-type control instruction (only for ``equal to'') without sample filtering.} \label{Fig:curves}
\end{center}
\end{figure*}
\begin{figure*}[thb]
\begin{center}
 \includegraphics[width=1.0\linewidth]{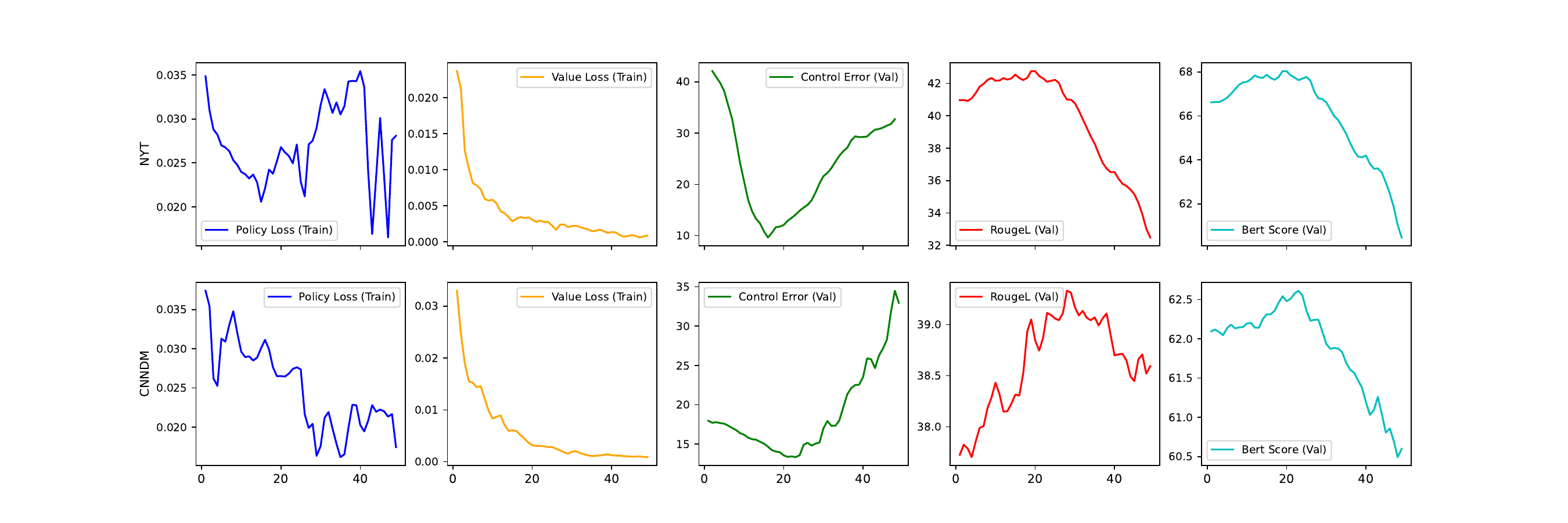}
 \caption{The Diagram of Learning Curves with GPT-S for multi-type control instructions without sample filtering.} \label{Fig:curves_mu_1}
\end{center}
\end{figure*}
\begin{figure*}[thb]
\begin{center}
 \includegraphics[width=1.0\linewidth]{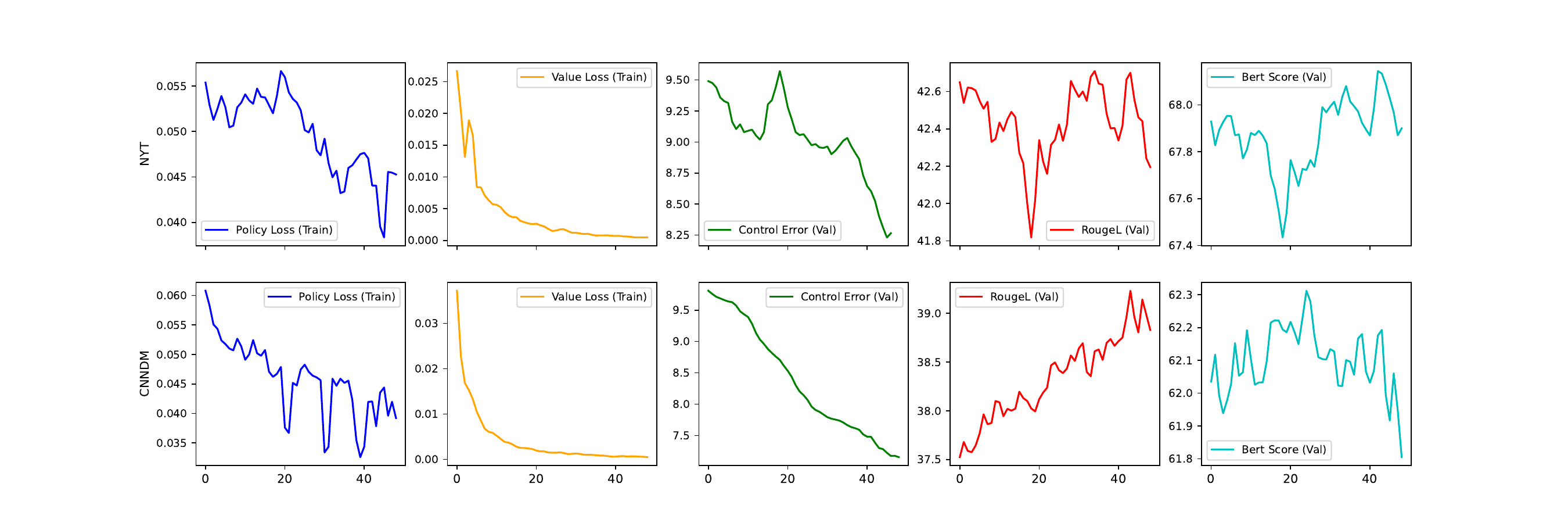}
 \caption{The Diagram of Learning Curves with GPT-S for single-type control instruction (only for ``equal to'') with sample filtering.} \label{Fig:curves_sg_8}
\end{center}
\end{figure*}
\begin{figure*}[thb]
\begin{center}
 \includegraphics[width=1.0\linewidth]{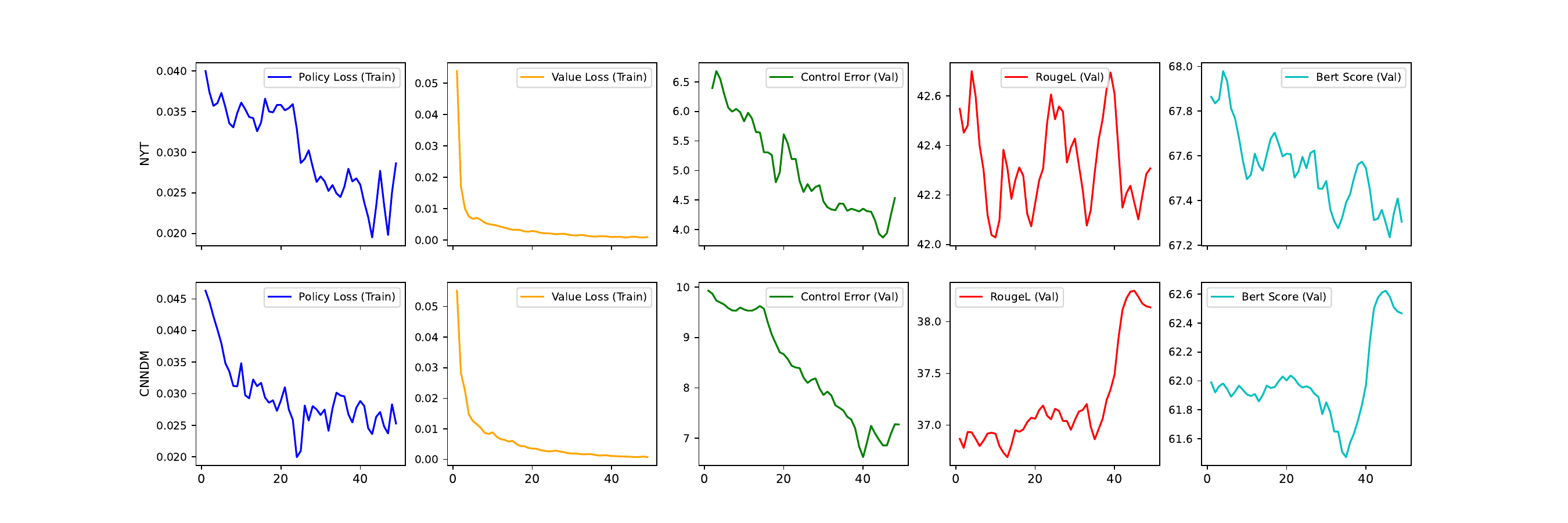}
 \caption{The Diagram of Learning Curves with GPT-S for multi-type control instructions with sample filtering.} \label{Fig:curves_mu_8}
\end{center}
\end{figure*}

To analyze the learning behavior, we visualize the learning curves of the policy loss and value loss on training set, control error and BERTscore (F1, in proportion) on validation set for a range of validation step. The results are first generated by small GPT-2 model on both NYT and CNNDM for single-type control (with only one control instruction which is ``equal to''), which are shown in Figure~\ref{Fig:curves}. We can see that as the decrease of policy loss and value loss, the validation reward increases relatively smoothly, while there is no clear decreasing trend of validation BERTscore. The indicates that even with small GPT-2 model, the relevance can be preserved as the control accuracy increase during the RL finetuning. Figure~\ref{Fig:curves_mu_1} shows the corresponding learning curves of RL-finetuning with GPT-S for the case with multiple control types, where all the four control types in Table~\ref{tab:1} are equally sampled. We can see that in general, the value loss decreases smoothly, while the policy loss may fluctuate but with a decreasing trend in general. In terms of the validation control errors, the curve first decrease and then increase after a certain point. Also, we find that the corresponding RougeL curves and Bertscore Curves show the reverse behaviors in general. This indicate that under certain settings, higher control accuracy (lower control error) is associated with higher relevance metrics. Meanwhile, it is necessary to do early stopping or other regularization approaches to prevent over-fitting. Figure ~\ref{Fig:curves_sg_8} and Figure~\ref{Fig:curves_mu_8} show the corresponding learning curves of GPT-S for single-type and multiple-type control \textbf{with sample filtering}. We can see that the curves of policy losses in training seems to be smoother than the case without sample filtering. The validation control errors still decrease during the RL fine-tuning. Meanwhile, there is no clear trend of a decrease of RougeL and Bertscores as length control errors decreases. 

From all of these figures, we do not observe clear trade-off between the goodness of Rouge scores/Bertscore and the length control errors. This means our method can achieve better control accuracy without losing the quality of generated summaries in terms of major automatic metrics.

\end{document}